\pdfoutput=1

\documentclass[11pt]{article}

\usepackage{ACL2023}

\usepackage{siunitx}  
\usepackage{float}

\usepackage{times}
\usepackage{latexsym}
\usepackage[T1]{fontenc}
\usepackage{subcaption}
\usepackage[utf8]{inputenc}
\usepackage{mathtools}
\usepackage{microtype}
\usepackage{tabularx}     
\usepackage{booktabs}     

\usepackage{inconsolata}
\usepackage{graphicx}
\usepackage{caption}
\usepackage{subcaption}
\usepackage{tabularx,booktabs}

\usepackage{algorithm}
\usepackage{algorithmic}

\usepackage{booktabs}
\usepackage{graphicx}
\usepackage{amsmath}
\usepackage{amssymb}
\usepackage{amsfonts}
\usepackage{algorithmic}
\usepackage{algorithm}
\usepackage{bbm}
\usepackage{multirow}
 \usepackage{listings}
 
\usepackage{booktabs}
\usepackage{afterpage}

\usepackage{xtab}
\usepackage{longtable}

\title{Zero-RAG: Towards Retrieval-Augmented Generation with Zero Redundant Knowledge}

\author{
  Qi Luo$^{1,*}$ \quad
  Xiaonan Li$^{1,*}$ \quad
  Junqi Dai$^{1}$ \quad
  Shuang Cheng$^{1}$ \quad
  Yining Zheng$^{1}$ \quad
  Xipeng Qiu$^{1,2,\dagger}$\\[4pt]
  $^1$School of Computer Science, Fudan University \\
  $^2$Shanghai Innovation Institute\\
  \texttt{qiluo22@m.fudan.edu.cn}
}

\begin{document}
\pagestyle{plain} 
\maketitle

\begin{abstract}
Retrieval-Augmented Generation has shown remarkable results to address Large Language Models' hallucinations, which usually uses a large external corpus to supplement knowledge to LLMs. However, with the development of LLMs, the internal knowledge of LLMs has expanded significantly, thus causing significant knowledge redundancy between the external corpus and LLMs. On the one hand, the indexing cost of dense retrieval is highly related to the corpus size and thus significant redundant knowledge intensifies the dense retrieval's workload. On the other hand, the redundant knowledge in the external corpus is not helpful to LLMs and our exploratory analysis shows that it instead hurts the RAG performance on those questions which the LLM can answer by itself. To address these issues, we propose Zero-RAG to tackle these challenges. Specifically, we first propose the Mastery-Score metric to identify redundant knowledge in the RAG corpus to prune it. After pruning, answers to "mastered" questions rely primarily on internal knowledge of the LLM. To better harness the internal capacity, we propose  Query Router and Noise-Tolerant Tuning to avoid the irrelevant documents' distraction and thus further improve the LLM's utilization of internal knowledge with pruned corpus. Experimental results show that Zero-RAG prunes the Wikipedia corpus by 30\% and accelerates the retrieval stage by 22\%, without compromising RAG's performance.

\end{abstract}

\begin{figure}[!t]
  \centering
  \begin{subfigure}[t]{0.35\textwidth}
    \centering
    \includegraphics[width=\textwidth]{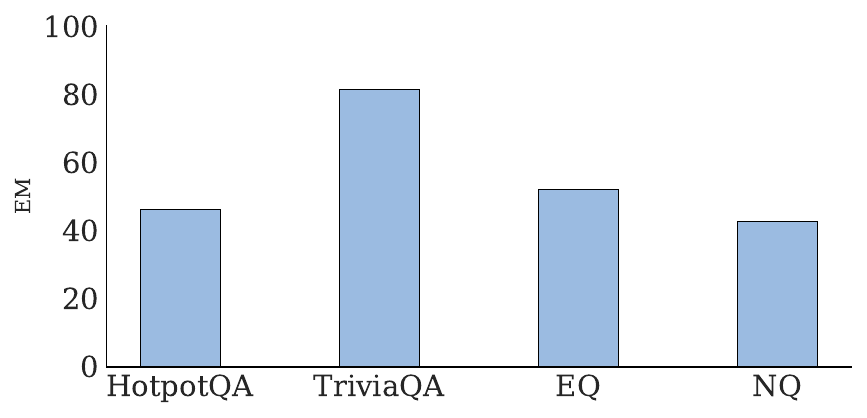}

    \caption{Exact Match (EM) recall on four Wikipedia-based QA datasets, evaluated using Llama3.3-70B, shows a substantial overlap between the model’s knowledge and Wikipedia.}

    \label{fig:sub2_Knowledge_overlap}
  \end{subfigure}
  \hfill
  \begin{subfigure}[t]{0.35\textwidth}
    \centering
    \includegraphics[width=\textwidth]{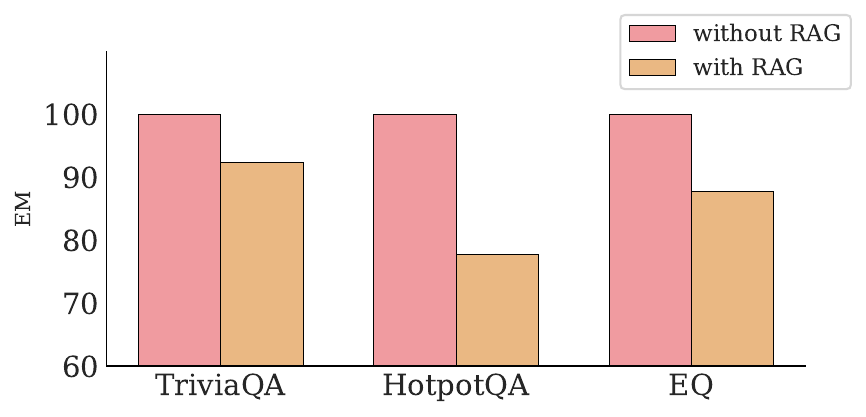}
    \caption{Comparison of performance on previously mastered questions: without versus with redundant knowledge retrieval.}
    \label{fig:sub3}
  \end{subfigure}
\caption{Overview of knowledge redundancy: (a) shows the overlap between LLM and corpus knowledge; (b) indicates that a large part of LLM knowledge is derived from Wikipedia; (c) reveals that redundant knowledge degrades performance on originally correct responses.}
\label{fig:three_at_top}
\end{figure}

\section{Introduction}

Retrieval-Augmented Generation (RAG) has become a important research topic to address Large Language Models' (LLMs) hallucination issues\cite{
lin-etal-2022-truthfulqa,
wang2023surveyfactualitylargelanguage,
gao2024retrievalaugmentedgenerationlargelanguage, 
fan2024surveyragmeetingllms, 
li2024llatrievalllmverifiedretrievalverifiable,
yu2024evaluationretrievalaugmentedgenerationsurvey}.
By integrating the internal knowledge of LLMs with a comprehensive range of external information, RAG enables LLMs to significantly enhance their factual accuracy while generating high-quality, informative responses~\citep{Self-RAG,li2024llatrievalllmverifiedretrievalverifiable}.

RAG usually integrates the LLM with a large external corpus~\citep{ratro}, which leads to a high heavy overload. With the development of LLMs, \citep{grattafiori2024llama3herdmodels,
deepseekai2025deepseekr1incentivizingreasoningcapability,
geminiteam2024geminifamilyhighlycapable} the internal knowledge of LLMs increases. For example, existing research has shown that the  LLM's knowledge density doubles every 100 days~\citep{densing_law}.
And thus more and more knowledge in the external corpus becomes redundant.
As shown in Figure~\ref{fig:sub2_Knowledge_overlap}, Llama3.3-70B can achieve at least 40\% accuracy on Wikipedia-related questions, which shows significant redundancy between the external corpus and the LLM's knowledge. 

In this paper, we focus on addressing the challenges of redundant knowledge between the external corpus and the LLM. 
On the one hand, the encoding and indexing workload of dense retrieval is highly related to the corpus size, and thus massive redundant knowledge intensify the dense retrieval's workload. On the other hand, the redundant knowledge in the external corpus is not helpful to the LLM, since the LLM contain that knowledge in parameters. Additionally, our exploratory experiment shows that the redundant knowledge even hurts the performance. Specifically, we evaluate the LLM’s performance on questions it has mastered when redundant knowledge is included. We filter questions that the LLM can correctly answer by itself and observe the accuracy when the corresponding passage is added to its context.
The results are shown in Figure~\ref{fig:sub3}.
We find adding the redundant knowledge to the LLM instead degrades its performance by about 20 points. Although the corresponding knowledge is supplemented to the LLM, it instead hurts the LLM's performance on those mastered questions. This shows that the redundant knowledge may distract the LLM and hinder it from utilizing corresponding knowledge.

To address these issues, we propose \textbf{Zero-RAG} to reduce knowledge redundancy between the corpus and the LLM and thus prune the redundant knowledge without compromising RAG's performance. Specifically, we propose \textbf{Mastery-Score} to measure how well the LLM masters a corpus passage and to prune those passages with high Mastery-Scores. In this way, we can significantly reduce the corpus size and retain the necessary knowledge in the corpus. 
After the corpus pruning, answering questions already mastered by the LLM, relies primarily on their internal knowledge. Therefore, we further propose the following two modules to help the LLM better utilize its internal knowledge under pruned corpus: \textbf{Query Router:} it first determines whether the LLM can answer the question by itself and let the LLM directly answer those mastered questions, which can help the LLM avoid the distraction of irrelevant documents and better utilize its internal knowledge; \textbf{Noise-Tolerant Tuning:} this makes LLM more robust and correctly utilize internal knowledge when the irrelevant documents are retrieved and put to its context. Through the synergy between the query router and noise-robust tuning, Zero-RAG can make the LLM better utilize its internal knowledge when the external corpus is significantly pruned. We summarize our contributions as follows:
\begin{itemize}
    \item To the best of our knowledge, Zero-RAG is the first to explore RAG-oriented corpus pruning by removing the knowledge redundancy between the corpus and the LLM, without compromising RAG's performance
    \item Experimental results on four Fact-intensive QA datasets show that Zero-RAG can significantly prune 30\% of the external corpus, accelerate the retrieval by 22\% and retain the RAG's performance.
    \item We will release the code, model, and pruned corpus. We hope that Zero-RAG can inspire researchers of the important design choices about pruning RAG corpus and pave the way for further improvements.
\end{itemize}

\section{Task Definition}

\begin{figure}
    \centering
    \includegraphics[width=0.7\linewidth]{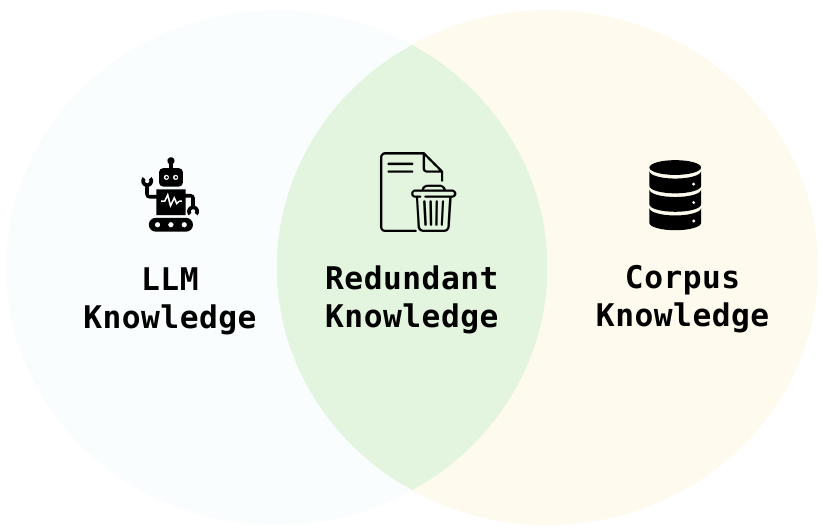}
    \caption{Relationship between model‐internal knowledge and corpus knowledge.  
The left circle denotes the knowledge already encoded in the language model ($\mathcal{K}_{\mathcal{M}}$), while the right circle denotes the out-of-model corpus $\mathcal{D}$.  
Their intersection, $\mathcal{D}_{\text{redundant}}$, contains facts duplicated in both sources and is therefore a prime target for pruning during \emph{Zero-RAG}}
    \label{fig:redundancy-venn}
\end{figure}

We formally define the problem of redundant knowledge in Retrieval-Augmented Generation (RAG) as follows. Given an external corpus $\mathcal{D} = \{d_i\}_{i=1}^{|\mathcal{D}|}$ and an LLM whose internal knowledge is encoded within its parameters, knowledge redundancy occurs when passages in $\mathcal{D}$ overlap significantly with the knowledge mastered by the LLM(see Figure~\ref{fig:redundancy-venn}). Such redundancy increases retrieval costs and may degrade LLM performance by introducing distracting or unnecessary context.

Our goal is thus to identify and prune redundant knowledge to produce a concise corpus $\mathcal{D}_{\text{retained}} \subset \mathcal{D}$, minimizing retrieval overhead without substantially compromising RAG effectiveness:

\begin{equation}
\mathcal{D}_{\text{redundant}} = \{d_i \in \mathcal{D} \mid \text{Overlap}(d_i, \mathcal{K}_{\mathcal{M}}) \leq \tau\},
\end{equation}

\begin{equation}
\mathcal{D}_{\text{retained}} = \mathcal{D} \setminus \mathcal{D}_{\text{redundant}}.
\end{equation}

where $\text{Overlap}(d_i, \mathcal{K}_{\mathcal{M}})$ measures the degree of overlap between document $d_i$ and the model's internal knowledge $\mathcal{K}_{\mathcal{M}}$.

\section{Zero-RAG}
\label{sec:zerorag}
RAG usually augments LLMs' knowledge by supplementing it with information from an external corpus. With the development of LLMs~\citep{densing_law}, the LLM's internal knowledge has increased significantly and thus causes significant knowledge redundancy between the RAG corpus and LLMs, which not only intensifies the retrieval's workload but also degrades the RAG performance (see Figure~\ref{fig:sub3}). In this section, we propose Zero-RAG to address these issues. In Zero-RAG, we first propose Mastery-Score to measure how well the LLM masters a passage from the corpus with high Mastery-Scores. After the corpus pruning, answering mastered questions mainly relies on LLM's internal knowledge. Therefore, we further propose Query Router and Noise-Tolerant Tuning to help the LLM better utilize its internal knowledge with the pruned corpus. We introduce these modules and the overall pipeline as follows.

 \begin{figure*}[t]
  \centering
  \includegraphics[width=0.9\textwidth]
  {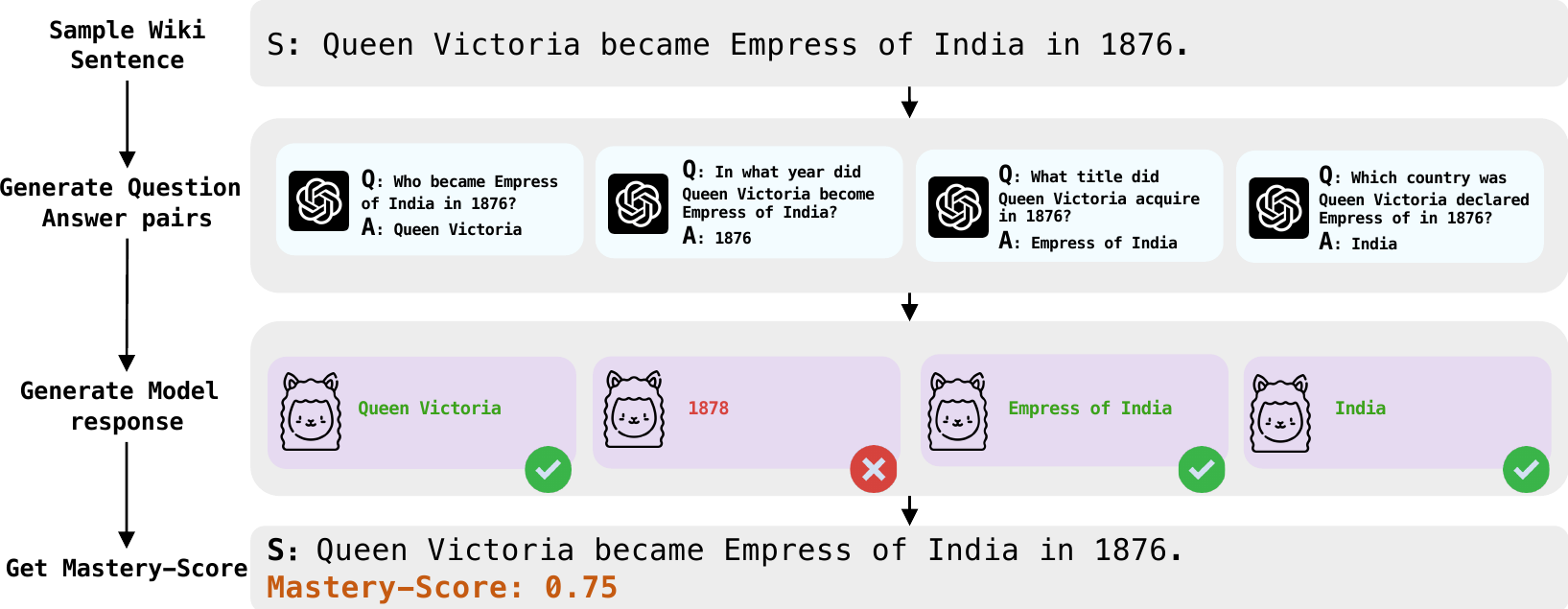}
\caption{Mastery-Score Construction Pipeline: For a given sentence, the mastery score is computed in four steps: (1) sampling candidate sentences from Wikipedia; (2) generating $n$ QA pairs for each sentence; (3) evaluating LLM performance on these QA pairs; and (4) calculating the final score based on the LLM's responses.}
  \label{Knowledge_score}
\end{figure*}

\subsection{Mastery-Score}


Existing works usually use loss-based strategies to probe whether the LLM memorizes a specific piece of knowledge\cite{fang2024wrongperplexitylongcontextlanguage, chen2024longloraefficientfinetuninglongcontext,ding2023longnetscalingtransformers1000000000, peng2023yarnefficientcontextwindow}. However, memorizing the knowledge doesn't ensure mastering it. \citet{llm_physics_31} finds that when the LLM memorizes a specific knowledge, it may not be able to use it after Supervised Fine-Tuning (SFT). To address this issue, we propose master-score to identify whether the LLM can flexibly utilize the knowledge after SFT. Given each passage's Mastery-Score, we can prune those documents with high Master-Score to reduce the knowledge redundancy between RAG corpus and LLM.


 \begin{figure*}[t]
  \centering
  \includegraphics[width=0.9\textwidth]
  {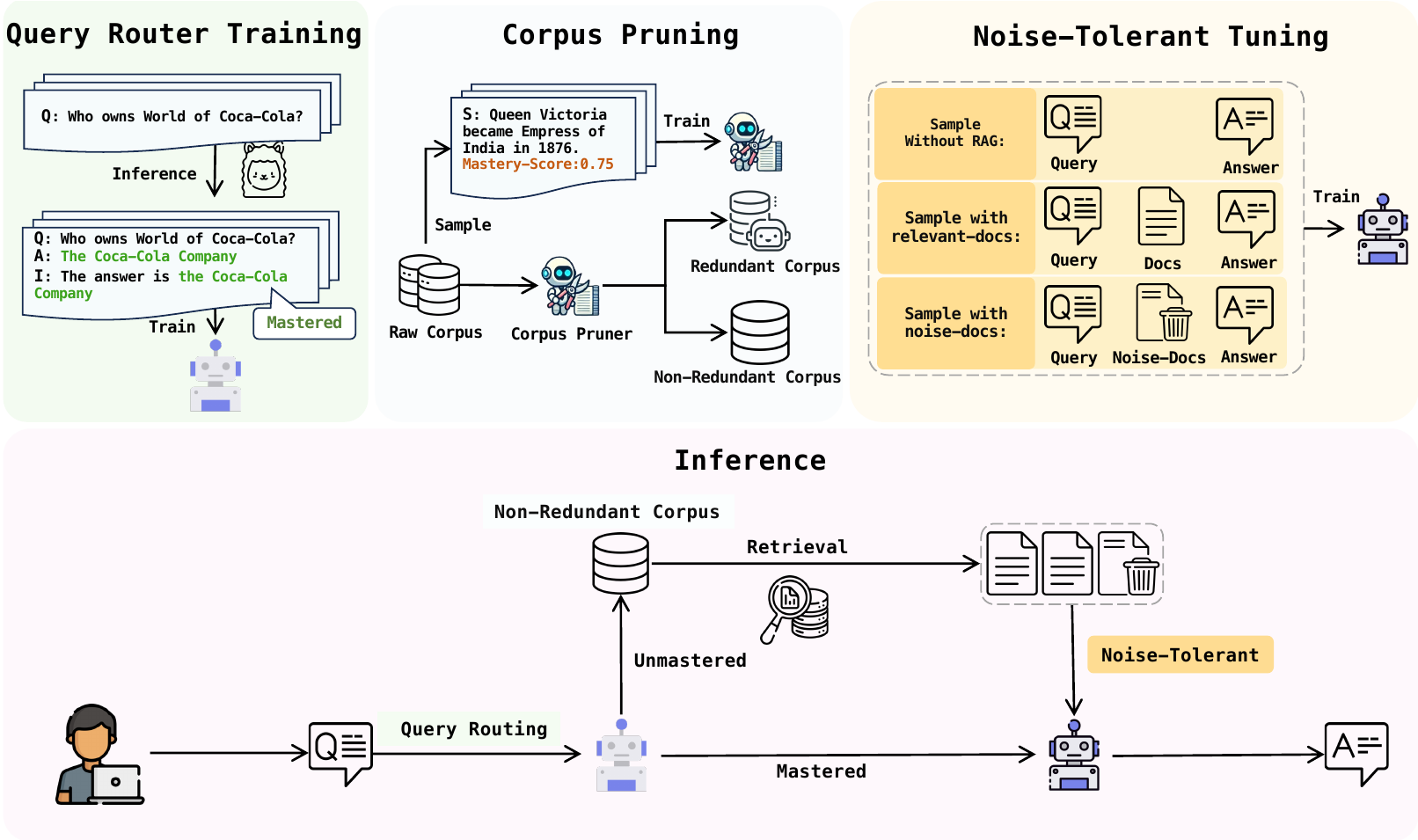}
    \caption{Overall architecture of \textsc{Zero-RAG}. The pipeline comprises four key stages:
    (1) \emph{Corpus Pruning via Mastery-Score}, which filters out already-mastered documents;
    (2) \emph{Query Router}, which dynamically decides whether to retrieve or not;
    (3) \emph{Noise-Tolerant Tuning}, ensuring robustness against partially relevant or irrelevant documents;
    and (4) the final \emph{Inference} stage that integrates all previous steps.}
  \vspace{-15pt}
  \label{fig_demonstration_retrieval_first_page}
\end{figure*}

We illustrate the process of computing the Mastery-Score in Figure~\ref{Knowledge_score}. For any sentence \( s \), we calculate its Mastery-Score \( M(s) \) in two steps:

    \paragraph{Construct QA Pairs:}  
    Use an LLM to generate \( n \) question-answer pairs \(\{(q_i, a_i)\}_{i=1}^n\) from \( s \). The number \( n \) is chosen based on the complexity of \( s \), ensuring that each question \( q_i \) can be answered directly using \( s \).
    
    \paragraph{Compute the Score:}  
    Calculate the average Exact Match (EM) score over the \( n \) pairs:
    
    \begin{small}
    \begin{align}
    M(s) = \frac{1}{n} \sum_{i=1}^n \mathrm{EM}\bigl(a_i, L(q_i)\bigr),
    \end{align}
    \end{small}

    where \( L(q_i) \) is the answer generated by the LLM for question \( q_i \), \(\mathrm{EM}(a_i, L(q_i))\) is the Exact Match score.

Since the mastery-score measures how well the LLM can answer questions related to the knowledge, it naturally reflects how well the LLM can utilize the knowledge after SFT.

    
    
    
    
    


\paragraph{Corpus Pruner Learning:} 
Measuring the mastery-score on one sentence may require several times LLM inference and directly applying it to the whole RAG corpus requires infeasible computational resources.
We thus propose a mastery-classifier to address this issue.
The core idea is using a neural classifier to predict the Mastery-Score for each sentence, which quantifies how well the language model \( L \) masters the information in sentence \( s_i \). The score is normalized to the range \([0,1]\), where 0 indicates complete redundancy and 1 indicates critical information.

Due to the high cost of annotating every sentence in the Wiki dataset, we design a regression model \( f_\theta: s_i \mapsto \hat{m}_i \) to predict the Mastery-Score.

We construct a training set \(\mathcal{D} = \{(s_i, m_i)\}_{i=1}^N\), where \( m_i \) is the ground truth Mastery-Score for sentence \( s_i \). The regression model is trained by minimizing the Mean Squared Error (MSE):

\begin{small}
\begin{align} 
\mathcal{L}(\theta) = \frac{1}{N} \sum_{i=1}^N \Bigl( f_\theta(s_i) - m_i \Bigr)^2,
\end{align}
\end{small}

where \( f_\theta(s_i) \) is the predicted Mastery-Score for \( s_i \), \( m_i \) is the true Mastery-Score for \( s_i \), \( \theta \) denotes the model parameters.

The output \( \hat{m}_i = f_\theta(s_i) \) represents the probability that sentence \( s_i \) is redundant. We prune a sentence when \( \hat{m}_i < \tau \) (a preset threshold). The complete corpus pruning procedure is detailed in Appendix~\ref{corpus_prune_prediction}.






\subsection{Query Router}

Even with pruning, some queries may still trigger retrieval of irrelevant documents—particularly if the LLM itself can answer a query without external context. In such cases, retrieval not only brings noise but also increases the overall latency. Therefore, Zero-RAG employs a \emph{query router} to determine whether a query requires additional information from the database. If the LLM can confidently answer using its internal knowledge alone, the query bypasses the retriever entirely. By selectively routing only queries that need external context, we reduce both computational overhead and the risk of retrieving noisy, off-topic documents. Next, we will explain how to train the Query Router by discussing both data construction and the learning process.

\textbf{Data Collection for query router} For a downstream dataset, we first evaluate the dataset using the Noise-Tolerant model to determine the model’s performance with various queries. Based on the model’s performance, we assign a classification label—either \textit{mastered} or \textit{unmastered}—to each data point. These labels represent whether the model is already knowledgeable about the corresponding question. Using this labeled dataset, we train a classifier to develop the Question Pruner. 

\textbf{Query router Learning} We train the query router using a binary classification approach, where each query is labeled to indicate whether the model already possesses sufficient knowledge to answer it. The loss function employed for training the query router is defined as:

\begin{small}

\begin{equation}
\mathcal{L}(\theta) = -\frac{1}{N} \sum_{i=1}^N y_i \log f_i,
\end{equation}
\end{small}

By accurately determining the familiarity of the model with the current query, the query router effectively prevents unnecessary access to \( \mathcal{D}_{\text{unmastered}} \).

\subsection{Noise-Tolerant Tuning}
\label{Noise-Tolerant-section}

Despite corpus pruning and query routing, there may be scenarios in which a query genuinely requires a pruned document or inadvertently retrieves unrelated content. To mitigate the adverse effects of such “noisy” documents, we adopt a \emph{noise-tolerant} fine-tuning scheme. Under this scheme, the model is explicitly trained to disregard or downweight irrelevant documents while relying on its internal knowledge. This further increases Zero-RAG’s robustness: even when the retrieved documents are partially or fully unrelated, the model can still produce accurate answers by leveraging its own learned information. Next, we explain how to implement Noise-Tolerant Tuning by detailing both the data construction and training processes.

\textbf{Data collection for Noise-Tolerant tuning} We design training samples in three formats:
\begin{itemize}
    \item Sample with relevant docs: Query \(q\) with relevant documents \(r_p\) and answer \(a\).
    \item Sample with noise docs: Query \(q\) with distracting documents \(r_n\) and answer \(a\).
    \item Sample without RAG: Query \(q\) with answer \(a\).
\end{itemize}

Our fine-tuning loss then unifies these formats as:

\begin{small}
\begin{equation}
\begin{aligned}
\mathcal{L} =\ &\underbrace{-\mathbb{E}\big[\log p_{\theta}(a|q)\big]}_{\mathclap{\text{Retrieval-Free}}} \\
             &-\underbrace{\mathbb{E}\big[\log p_{\theta}(a|q,r_p)\big]}_{\mathclap{\text{Retrieval-Augment}}} \\
             &-\underbrace{\mathbb{E}\big[\log p_{\theta}(a|q,r_n)\big]}_{\mathclap{\text{Noise-Suppression}}}
\end{aligned}
\end{equation}
\end{small}

\subsection{Zero-RAG Inference}
\label{subsec:Inference}
Our inference pipeline combines the trained Corpus Pruner, Query Router, and Retriever to efficiently generate answers while avoiding redundant information. The process is as follows:

\begin{enumerate}
    \item \textbf{Corpus Pruning:}  
    We initiate the inference process by pruning the original corpus \( D \) to remove redundant information. Utilizing the Corpus Pruner, we assign a Mastery-Score to each sentence and filter out samples with high scores, resulting in a non-redundant subset $D_{\text{non-red}}$. Further details are provided in Appendix~\ref{corpus_prune_prediction}.

    \item \textbf{Query  Routing:}  
    Then, for each query \( q \), the Query Router checks if the model already has sufficient internal knowledge to answer it.
    
    \item \textbf{Final Inference:}  
    For a \emph{mastered} query, the model answers directly using its internal knowledge. For an \emph{unmastered} query, the Retriever searches $D_{\text{non-red}}$ for a few relevant documents, which are provided to the model to generate the final answer.
\end{enumerate}








\begin{table*}[t]
\centering
\small
\setlength\tabcolsep{8pt}
\begin{tabular}{@{}lcccc@{}}
\toprule
20920\textbf{Method} & \textbf{PopQA} & \textbf{HotpotQA} & \textbf{TriviaQA} & \textbf{EntityQuestions} \\
\midrule
\multicolumn{5}{c}{\textit{Llama3-70B}} \\
\midrule
Llama3-70B-Instruct & 14.08 & 43.71  & 80.66 & 51.91 \\
~~+ Retrieval & 15.62  & 41.40 & 76.44 & 48.25
 \\
Noise-Tolerant Tuning & 25.21
& 42.70  & 81.43 &  54.14 \\
~~+ Retrieval & 39.36  & 49.67
 & 81.90 & 65.25\\
Zero-RAG (No Pruning) & 31.72 & 45.00  & 81.80 &  60.82 \\ \midrule

Zero-RAG (- 10\% Corpus) & 30.81 &43.84  & 81.50  & 58.92  \\
Zero-RAG (- 30\% Corpus) & 30.67 & 43.30 & 81.00

 & 57.82 \\ 
Zero-RAG (- 50\% Corpus) & 30.32 & 41.93 & 80.53 &  56.11  \\
Zero-RAG (- 70\% Corpus) & 29.48 & 40.69 & 80.40 &  55.41 \\ 
\midrule

\multicolumn{5}{c}{\textit{Llama3.3-70B}} \\
\midrule
Llama3.3-70B-Instruct & 16.25 & 46.20 & 81.43 & 52.01 \\
~~+ Retrieval & 16.35 & 43.35 & 79.32 & 50.71 \\
Noise-Tolerant Tuning & 32.77 & 47.50 & 82.19 & 55.38 \\
~~+ Retrieval & 38.94 & 49.12 & 81.50 & 65.16 \\
Zero-RAG (No Pruning) & 35.78 & 51.28 & 82.69 & 63.70 \\ \midrule
Zero-RAG (- 10\% Corpus) & 35.43 &  49.41 & 82.57 &  61.33 \\
Zero-RAG (- 30\% Corpus) & 34.80 & 48.52 & 82.42 & 60.43 \\
Zero-RAG (- 50\% Corpus) & 34.24 & 47.09 & 82.17  &  57.35 \\
Zero-RAG (- 70\% Corpus) & 32.91 & 46.20 & 82.07  & 56.29 \\

\bottomrule
\end{tabular}

\caption{Performance Comparison Across Datasets with Different Methods. Llama3-70B-Instruct methods are shown for comparison.}
\label{tab:main}

\end{table*}

\section{Experiment}

\subsection{Experimental Settings}
\textbf{Datasets} We evaluate on four standard QA benchmarks with distinct characteristics:
\textbf{EntityQuestions} \cite{sciavolino2021simple}: Wikidata-based simple entity queries for structured knowledge evaluation. \textbf{TriviaQA} \cite{joshi2017triviaqalargescaledistantly}: Large-scale evidence-based QA with document context requirements. \textbf{PopQA} \cite{wang2024webquestbenchmarkmultimodalqa}: Template-generated entity pairs (14k) with popularity annotations. \textbf{HotpotQA} \cite{yang2018hotpotqadatasetdiverseexplainable}: Multi-hop reasoning requiring cross-document analysis

\paragraph{Method Comparison} Since our work is the first to explore database pruning, a technique that can be combined with other RAG methods, so we compare our approach against two baselines: \textbf{Standard Instruct Models:} These are typical instruction-tuned models (e.g., llama3-70B-Instruct, llama3.3-70B-Instruct) that use the full database for both retrieval and answer generation, without any pruning. \textbf{Noise-Tolerant Models:} These models are fine-tuned using our Noise-Tolerant Tuning approach (see Section~\ref{Noise-Tolerant-section}), which incorporates database pruning into the training process to enhance retrieval efficiency and robustness.

\paragraph{Implementation Details} For our retrieval tasks, we use Wikipedia as the database due to its prominence in RAG research. We segment the Wikipedia text into sentences using Python’s NLTK sentence tokenizer. In total, the complete corpus was divided into 138,390,600 sentences. For each selected sentence, we generate corresponding question-answer pairs using \texttt{GPT-4o-mini}. We design a prompt template that instructs the model to create a question whose answer is contained in the sentence ,show in Appendix~\ref{prompt}. Depending on the sentence complexity, we generate \( n \) QA pairs per sentence. After generation, we filter out invalid QA pairs—those where the answer does not correctly reflect content in the sentence or does not meet our quality criteria—to ensure that only high-quality training data are used. During the corpus pruning phase, a 7B model is trained to predict the Mastery-Score (MS) of a 70B model. For the SFT training phase, we utilize LoRA for fine-tuning with a learning rate of \(3 \times 10^{-4}\). In the retrieval phase, we employ \texttt{stella\_en\_400M\_v5} as the retrieval model and set the number of document candidates per query to 20. \looseness=-1

\subsection{Main Results}
As illustrated in Table~\ref{tab:main}, we evaluate both Llama3-70B and Llama3.3-70B on four QA datasets. Compared to the baseline zero-RAG system that uses the full corpus, pruning 30\% of the database results in only minimal performance degradation—averaging less than a two-point drop at moderate pruning levels and around three points at a 70\% pruning ratio. This indicates that abundant redundant knowledge in the original corpus provides negligible benefit. Specifically, for TriviaQA, removing 70\% of the corpus results in merely a 0.62-point drop, suggesting that much of TriviaQA-related knowledge is redundant and that eliminating it has almost no adverse effect on performance.

Furthermore, for TriviaQA,EntityQuestion, HotpotQA, simply adding corpus for RAG leads to a slight drop in scores. However, once Noise-Tuning is applied, performance not only recovers but actually surpasses the original baseline, indicating that our approach effectively enhances RAG’s robustness and overall utility.

Additionally, under the same pruning ratios, Llama3.3-70B consistently outperforms Llama3-70B, implying that Llama3.3-70B possesses a larger reservoir of overlapping knowledge with Wikipedia.

\begin{table*}[]
\centering
\resizebox{\textwidth}{!}{%
\begin{tabular}{lcccccccccccc}
\toprule

\multirow{2}{*}{\textbf{Method}} & \multicolumn{3}{c}{\textbf{PopQA}} & \multicolumn{3}{c}{\textbf{HotpotQA}} & \multicolumn{3}{c}{\textbf{TriviaQA}} & \multicolumn{3}{c}{\textbf{EntityQuestions}} \\ 
\cmidrule(lr){2-4} \cmidrule(lr){5-7} \cmidrule(lr){8-10} \cmidrule(lr){11-13}

 & \textbf{Top5} & \textbf{Top10} & \textbf{Top20} & \textbf{Top5} & \textbf{Top10} & \textbf{Top20} & \textbf{Top5} & \textbf{Top10} & \textbf{Top20} & \textbf{Top5} & \textbf{Top10} & \textbf{Top20} \\ \hline
Llama3.3-70B-Instruct & 16.25 & 16.25 & 16.25 & 46.20 & 46.20 & 46.20 & 81.43 & 81.43 & 81.43 & 52.01 & 52.01 & 52.01 \\
~~+ Retrieval & 13.87 & 15.34 & 16.35 & 43.42 & 44.86 & 43.35 & 74.85 & 77.57 & 79.32 & 50.24 & 53.73 & 50.71 \\
Noise-Tolerant Tuning & 32.77 & 32.77 & 32.77 & 47.50 & 47.50 & 47.50 & 82.19 & 82.19 & 82.19 & 55.38 & 55.38 & 55.38 \\
~~+ Retrieval & 38.09 & 38.66 & 38.94 & 48.29 & 49.05 & 49.12 & 80.54 & 81.41 & 81.50 & 63.76 & 64.47 & 65.16 \\
Zero-RAG (No Pruning) & 35.17 & 35.73 & 35.78 & 50.98 & 51.16 & 51.28 & 82.46 & 82.58 & 82.69 & 62.76 & 63.28 & 63.70 \\ \midrule
Zero-RAG (- 30\% Corpus) & 33.87 & 33.99 & 34.01 & 48.01 & 48.50 & 48.52 & 82.31 & 82.41 & 82.42 & 59.75 & 60.08 & 60.43 \\ 
\bottomrule

\end{tabular}%
}

\caption{An ablation study assessing the performance of Zero-RAG on Llama3.3-70B under various top-$k$ settings (Top5, Top10, and Top20).}
\label{topk_ablation}

\end{table*}

\begin{table}[t]
\centering
\small
\setlength\tabcolsep{0.5pt}
\resizebox{0.48\textwidth}{!}{

\begin{tabular}{lcccc}
\toprule
\multirow{2}{*}{\textbf{Method}} & \multicolumn{2}{c}{\textbf{TriviaQA}} & \multicolumn{2}{c}{\textbf{HotpotQA}} \\ 
\cmidrule(lr){2-3} \cmidrule(lr){4-5}
 & \textbf{Prune Ratio} & \textbf{EM} & \textbf{Prune Ratio} & \textbf{EM} \\ 
\midrule
Llama3.3-70B-Instruct & - & 81.43 & - & 46.20 \\
~~+ Retrieval & 0\% & 79.32 & 0\% & 43.35 \\
\midrule
Zero-RAG & 30\% & 82.42 & 30\% & 48.52 \\
~~- Corpus Prune & 0\% & 82.69 & 0\% & 51.28 \\
~~- Query Router & 30\% & 81.50 & 30\% & 43.35 \\
~~- Noise-Tolerant Tuning & 30\% & 80.55 & 30\% & 42.82 \\
\bottomrule
\end{tabular}%
}

\caption{Ablation study assessing the effectiveness of Zero-RAG components on the TriviaQA and HotpotQA datasets using the Llama3.3-70B-Instruct model. Exact Match (EM) scores are reported for various pruning ratios.}

\label{tab:component_ablation}
\end{table}

\begin{table}[h!]
\centering

\resizebox{0.48\textwidth}{!}{%
\begin{tabular}{l
                S[table-format=2.2]
                S[table-format=2.2]
                S[table-format=2.2]
                S[table-format=2.2]}
\toprule
\textbf{Prune Ratio} & \textbf{HotpotQA} & \textbf{EntityQ} & \textbf{TriviaQA} & \textbf{Avg} \\
\midrule
0\%  & 14.65 & 10.96 & 11.24 & 12.28 \\
30\% & 10.89 &  9.09 &  9.00 &  9.66 \\
\bottomrule

\end{tabular}
}
\caption{Retrieval latency (in seconds) under different prune ratios for various datasets. 
The \textit{Avg} column represents the mean latency across all three datasets.}
\label{tab:latency_per_query}
\end{table}

\subsection{Analyses}
\textbf{The Effect of Different Components} 
We evaluate the effectiveness of each component on TriviaQA and HotpotQA. In Table~\ref{tab:component_ablation}, we present the performance changes that occur when each individual component is removed in isolation. First, we omit the \emph{Query Router}, which is designed to detect whether the LLM can confidently answer a query using its internal knowledge. This routing mechanism helps avoid unnecessary retrieval for “known” queries, thereby preventing the introduction of noisy or redundant documents. Removing the Query Router leads to a noticeable drop in performance, indicating that unnecessary retrieval can degrade both accuracy and efficiency. Next, we remove the \emph{Noise-Tolerant Tuning} module, which explicitly trains the model to disregard irrelevant or misleading documents when partial or fully unrelated content is retrieved. This approach is motivated by the potential for genuine query needs that still require pruned documents, or instances where retrieval may unintentionally return off-topic material. Disabling Noise-Tolerant Tuning yields a significant decrease in Exact Match (EM), underscoring its crucial role in preserving model robustness amidst noisy retrieval results. Finally, the \emph{Corpus Pruning} component effectively eliminates redundant and non-essential knowledge from the database. By maintaining a leaner yet sufficiently informative database, Zero-RAG can streamline retrieval without sacrificing accuracy. Even after a 30\% reduction in database size, the model continues to perform strongly—demonstrating the efficacy of pruning in reducing overhead while retaining critical information.

\begin{table}[t]
\centering
\large 
\setlength{\tabcolsep}{5pt} 
\renewcommand{\arraystretch}{1.1} 
\resizebox{0.48\textwidth}{!}{%
\begin{tabular}{@{}lcccc@{}}
\toprule
\textbf{Method} & \textbf{PopQA} & \textbf{HotpotQA} & \textbf{TriviaQA} & \textbf{EQ} \\
\midrule
Llama3-8B-Instruct   & 7.98  & 32.67 & 68.75 & 33.30 \\
\quad + Retrieval     & 11.84 & 35.42 & 71.23 & 45.84 \\
Noise-Tolerant Tuning & 12.96 & 32.57 & 68.74 & 39.84 \\
\quad + Retrieval     & 30.32 & 38.80 & 75.63 & 55.03 \\
Zero-RAG (No Pruning)             & 21.29 & 35.21 & 72.08 & 48.71 \\ \midrule
Zero-RAG (- 30\% Corpus)   & 18.91 & 31.89 & 67.93 & 43.19 \\ 
\bottomrule
\end{tabular}
}
\caption{Performance of Llama3-8B with a 30\% pruned corpus as evaluated on four QA datasets. These objective results demonstrate that pruning the corpus enhances both retrieval efficiency and overall performance.}
\label{tab:8b_pruning_comparison}
\end{table}

\paragraph{Effect of Different Model Sizes}
Table~\ref{tab:8b_pruning_comparison} presents results for Llama3-8B on PopQA, HotpotQA, TriviaQA, and EntityQuestions. Even though this is a smaller model, the table shows it still contains substantial redundant knowledge. When 30\% of the corpus is pruned, the Exact Match (EM) recall decreases by only about four percentage points, underscoring the limited utility of the pruned data. These findings indicate that redundancy persists not only in large-scale models but also in smaller ones like Llama3-8B.

\paragraph{The Impact of the Number of Retrieved Documents}
To evaluate the impact of different top-$k$ settings on the performance of Zero-RAG, we conducted an ablation study across four datasets: PopQA, HotpotQA, TriviaQA, and EntityQuestions, utilizing the Llama3.3-70B-Instruct model. Specifically, we examined three top-$k$ settings—Top5, Top10, and Top20—to understand how varying the number of retrieved sentences influences the model's Exact Match (EM) scores. The detailed results are presented in Table~\ref{topk_ablation}. The ablation study reveals that Zero-RAG maintains robust performance across different top-$k$ settings. While EM scores exhibit slight variations with changes in top-$k$, the overall trend indicates that Zero-RAG effectively leverages retrieved information without being overly sensitive to the exact value of $k$. This insensitivity to the filtering threshold hyper-parameter underscores Zero-RAG's robustness and flexibility, demonstrating its capability to maintain performance across a range of retrieval configurations.

\paragraph{Optimizing Indexing and Retrieval Efficiency via Database Pruning}
Table~\ref{tab:latency_per_query} summarizes the retrieval latency (in seconds) for HotpotQA, EntityQuestions, and TriviaQA under two pruning ratios: 0\% (no pruning) and 30\%. With no pruning, the latency ranges from 10.96\,s to 14.65\,s (averaging 12.28\,s). By pruning 30\% of the corpus, latency decreases across all datasets (from 9.00\,s to 10.89\,s), bringing the average latency down to 9.66\,s. These results indicate that reducing the database size by 30\% yields a marked speedup (over 20\% improvement on average), thus enhancing both indexing and retrieval efficiency.


\begin{table*}[!t]
\centering
\footnotesize 
\setlength{\tabcolsep}{4pt} 
\renewcommand{\arraystretch}{1.1} 

\begin{tabularx}{\textwidth}{@{}lX@{}}
\toprule
\textbf{Field} & \textbf{Content} \\
\midrule
\textbf{Wiki Sentence} & 
\emph{“Queen Victoria became Empress of India in 1876.”} \\[0.5em]
\textbf{Q\&A Pairs} &
\textbf{Q1:} Who became Empress of India in 1876?\\
& \textbf{A1:} Queen Victoria\\[0.3em]
& \textbf{Q2:} In what year did Queen Victoria become Empress of India?\\
& \textbf{A2:} 1876\\[0.3em]
& \textbf{Q3:} What title did Queen Victoria acquire in 1876?\\
& \textbf{A3:} Empress of India\\[0.3em]
& \textbf{Q4:} Which British monarch became Empress of India in 1876?\\
& \textbf{A4:} Queen Victoria\\[0.5em]
\textbf{Predictions} &
\textbf{P1:} The answer is Queen Victoria. She was proclaimed Empress of India in 1876.\\[0.3em]
& \textbf{P2:} Queen Victoria became the Empress of India in 1876.\\[0.3em]
& \textbf{P3:} The answer is: Empress of India.\\[0.3em]
& \textbf{P4:} The answer is Queen Victoria. She was proclaimed Empress of India by the Royal Titles Act 1876.\\[0.5em]
\textbf{Eval Results} &
[Mastered, Mastered, Mastered, Mastered]\\[0.5em]
\textbf{Mastery-Score} &
1.0\\
\bottomrule
\end{tabularx}

\caption{A case study on the sentence 
“Queen Victoria became Empress of India in 1876,” demonstrating that the LLM had already mastered this content.
}
\label{tab:case_study_victoria}
\end{table*}

\paragraph{Case Study}
To further demonstrate the effectiveness of our corpus pruning strategy, we sampled a sentence from the pruned portion of the Wikipedia database and constructed corresponding question--answer pairs. Specifically, we chose the sentence \emph{“Queen Victoria became Empress of India in 1876.”}---which had been removed due to its high Mastery-Score---and derived four questions based on it. As shown in Table~\ref{tab:case_study_victoria}, Llama3-70B-Instruct correctly answered all of these questions, achieving an accuracy of 100\%. This result indicates that our pruning process indeed discarded knowledge the model had already “mastered”.

\section{Related Work}
\paragraph{Effective RAG}
A variety of methods have been proposed to improve the effectiveness of Retrieval-Augmented Generation (RAG). A more capable retriever supplies highly relevant context, reducing the language model’s workload and boosting overall performance. \citet{zhu2024longembedextendingembeddingmodels} describe a training-free approach that extends the retriever’s context window, allowing it to handle longer documents more effectively, while \citet{mao2024chatretrieveradaptinglargelanguage} adapt large language models (LLMs) for conversational retrieval, improving the relevance of information in dialogue settings. On the model side, recent work has focused on fine-tuning strategies for handling retrieved evidence more robustly. For example, \citet{zhang2024raftadaptinglanguagemodel} incorporate negative samples during training, enabling the model to deal better with irrelevant or noisy documents. Additionally, Chain-of-Note \cite{Yu2023ChainofNoteER} teaches the model to rely on its internal knowledge when irrelevant documents are retrieved, whereas SEAKR \cite{Yao2024SeaKRSK} uses self-aware uncertainty to re-rank candidate documents and select those that reduce the model’s uncertainty the most.


\paragraph{Efficient RAG}
\citet{fei2023extendingcontextwindowlarge} focuse on compressing the original prompt into a shorter, semantically equivalent representation. AutoCompressors\citet{chevalier2023adaptinglanguagemodelscompress}, ICAE\cite{ge2024incontextautoencodercontextcompression}, and RECOMP \cite{xu2023recompimprovingretrievalaugmentedlms}, explore methods to compress and represent long texts into summary vectors, ensuring compatibility with language models’ input constraints. \citet{Zhang2024AcceleratingRL} improves the efficiency of iterative retrieval by caching recently retrieved documents. ColPali \cite{faysse2024colpaliefficientdocumentretrieval} adapted Vision-Language Models (VLMs) for the retrieval task, simplifying the document retrieval pipeline to achieve efficient document retrieval. \citet{jiang2023activeretrievalaugmentedgeneration} and \citet{Liu2024RAISFLT} propose methods to determine whether retrieval is necessary based on the given question. Additionally, \citet{jeong2024adaptiveraglearningadaptretrievalaugmented} introduces strategies that adapt retrieval approaches according to the complexity of the question. \cite{zhuang2024efficientragefficientretrievermultihop} through iteratively generates new queries while circumventing the need for large language models

\section{Conclusion}

In this paper, we propose Zero-RAG to reduce the knowledge dedundancy between the RAG corpus and LLM. Specifically, we propose mastery-score to identify redundant knowledge in RAG corpus and prune it. Additionally, we prpose query router and noise-tolerant training to further improve RAG's utilization of LLM internal knowledge under the pruned corpus.
Experimental results on four Fact-intensive QA datasets show that Zero-RAG can significantly prune 30\% of the external corpus, accelerate the retrieval by 22\% and retain the RAG's performance. 


\clearpage  

\section*{Limitations}
we summarized our limitations as follows:
\begin{itemize}
    \item \textbf{Generalizability Across Domains}: Our study exclusively focuses on corpus pruning within Wiki-based databases. The effectiveness of Zero-RAG on other types of databases, such as domain-specific or multimodal datasets, has not been evaluated and remains to be explored.
    \item \textbf{Dependency on Initial Data Quality}: The performance of Zero-RAG is contingent upon the quality of the initial dataset. If the original data contains high levels of noise or inconsistencies, the pruning mechanism may inadvertently remove valuable information, potentially impacting model performance more significantly.
\end{itemize}
\clearpage  
\bibliography{references} 
\clearpage  
\appendix

\section{Details of Mastery-Score}

\subsection{Corpus Pruning}
\label{corpus_prune_prediction}

A sentence-based Mastery-Score is defined as:
\[
R_{\text{LM}}(s) = f_{\text{reg}}(s),
\]
where \(R_{\text{LM}}(s)\) denotes the sentence-level Mastery-Score predicted by the regression model \(f_{\text{reg}}\).

To quantify each sentence's contribution to the knowledge database, we compute \(M(s)\) for a sentence \(s\). A high Mastery-Score indicates that the sentence's knowledge is already well-represented (and thus redundant) within the language model, whereas a low Mastery-Score suggests unique or under-represented content. Hence, the Mastery-Score serves as a crucial criterion for determining whether a sentence should be pruned or retained.

However, effectively utilizing the Mastery-Score requires a clear threshold to distinguish redundant from unique sentences. Setting a fixed threshold \(\tau\) is challenging, as redundancy and informativeness can vary widely across datasets and LLMs. To address this, we adopt a dynamic, data-driven approach for determining \(\tau\), ensuring adaptability to different scenarios.

In this approach, we rank all sentences by their Mastery-Scores and compute a pruning threshold \(\tau\) based on a specified percentile \(p\). Formally,
\[
\tau = \mathrm{Percentile}(R_{\text{LM}}(s), p),
\]
where \(R_{\text{LM}}(s)\) is the Mastery-Score for each sentence in the database, and \(p\) is the pruning ratio dictating what fraction of sentences to retain. We then remove sentences whose scores exceed \(\tau\). The pruned database is:
\[
\mathcal{D}_{\text{mastered}} = \{\,s \in \mathcal{D} \mid R_{\text{LM}}(s) < \tau \},
\]
where \(\mathcal{D}\) is the original database and \(\mathcal{D}_{\text{mastered}}\) is the resulting subset after pruning.

By employing this dynamic threshold, we balance the retention of essential knowledge and the removal of redundant information. This not only improves retrieval efficiency but also ensures that the remaining database is more relevant for the target task.

During inference, the pruned database \(\mathcal{D}_{\text{pruned}}\) is queried to locate sentences that aid the language model (LLM) in answering a test question \(q_{\text{test}}\). However, retrieving sentences for questions the LLM already knows can introduce irrelevant documents and potentially degrade performance.

\subsection{Prompt}
\label{prompt}
To generate questions from a sentence, we use the following prompt:

\begin{lstlisting}[frame=none]
Instruction: Construct specific and complete questions based on the entities mentioned in the sentence. Ensure each question is self-contained and avoids pronoun references.

Reference: {reference}

Question 1:
Question 2:
Question 3:
...

\end{lstlisting}

To answer a given question, we use this prompt:

\begin{lstlisting}[frame=none]
Instruction: Answer the following question using as few words as possible (no more than 5 words). Ensure your response is brief and to the point.

Question: {question}

Answer:
\end{lstlisting}

For a given sentence, question, and answer triple, we use the following prompt to evaluate the quality of the generated QA pair:

\begin{lstlisting}[frame=none]

Instruction: Determine if the following question can be answered by the given reference. Respond with "yes" or "no".

Question: {question}
Reference: {reference}
Answer: {answer}
Response:

\end{lstlisting}

\section{Time Consumption}

We evaluated the inference time under different pruning ratios on three datasets: HotpotQA, EntityQuestion, and TriviaQA. As shown in Table~\ref{tab:prune_results}, Zero-RAG with a 30\% pruning ratio significantly reduces the overall time cost. For example, on the TriviaQA dataset, the total processing time decreases from about 1600\,s to around 400\,s. This improvement can be mainly attributed to two factors: (1) reduced indexing time due to a smaller index size, and (2) reduced retrieval overhead for questions that are pruned, thereby saving subsequent computation time. Moreover, as future models and knowledge bases continue to expand, the amount of redundant information that can be pruned is expected to increase, leading to even greater reductions in overall time consumption.

\begin{table}[t]
\centering
\small
\setlength\tabcolsep{8pt}

\captionsetup{width=0.48\textwidth} 

\begin{tabular}{lcc}
\toprule
\textbf{Dataset} & \textbf{Prune Ratio} & \textbf{Total Time (s)} \\ \midrule
\multirow{2}{*}{\textbf{HotpotQA}} & 0 & 717.13 \\
 & 30 & 531.94 \\ \midrule
\multirow{2}{*}{\textbf{EntityQuestion}} & 0 & 1924.47 \\
 & 30 & 1476.23 \\ \midrule
\multirow{2}{*}{\textbf{TriviaQA}} & 0 & 1655.54 \\
 & 30 & 441.80 \\ \bottomrule
\end{tabular}%

\caption{Performance Metrics under Different Prune Ratios for Various Datasets.}
\label{tab:prune_results}
\end{table}

\section{The effect of Differen Model}

 Table~\ref{tab:prune_results} reports the performance of the Qwen-2-72B model on PopQA, HotpotQA, TriviaQA. Compared to the baseline zero-RAG system that uses the full corpus, pruning 30\% of the database results in only minimal performance degradation—averaging less than a three-point drop at moderate pruning levels.

\begin{table}[t]
\centering
\small

\setlength\tabcolsep{8pt}
\resizebox{0.48\textwidth}{!}{%

\begin{tabular}{@{}lccc@{}}
\toprule
\textbf{Method} & \textbf{PopQA} & \textbf{HotpotQA} & \textbf{TriviaQA}  \\
\midrule
\multicolumn{4}{c}{\textit{Qwen-2-72B}} \\
\midrule
Llama3-70B-Instruct & 11.06 & 46.32 & 78.80   \\
~~+ Retrieval & 14.36 & 48.18 & 79.43  \\
Noise-Tolerant Tuning & 32.77 & 47.50 & 82.19   \\
~~+ Retrieval & 32.56 & 46.58 & 78.91   \\
Zero-RAG & 32.76 & 49.18 & 81.76   \\ \midrule
~~Zero-RAG (- 30\% Corpus & 29.13 & 45.17 & 80.90   \\
\bottomrule
\end{tabular}
}
\caption{Qwen-2-72B results}
\label{tab:pruning_comparison}
\end{table}


\end{document}